%% file: main.tex
\documentclass[sigconf]{acmart}

\AtBeginDocument{%
	\providecommand\BibTeX{{%
			\normalfont B\kern-0.5em{\scshape i\kern-0.25em b}\kern-0.8em\TeX}}}

\usepackage{epsfig}
\usepackage{graphicx}
\usepackage{amssymb}
\usepackage{subfigure}
\usepackage{comment}
\usepackage{xspace}

\usepackage{hyperref}
\hypersetup{
	colorlinks=true,
	linkcolor=blue,
	urlcolor=blue,
}

\copyrightyear{2020}
\acmYear{2020}
\setcopyright{acmcopyright}\acmConference[MM '20]{Proceedings of the 28th ACM
	International Conference on Multimedia}{October 12--16, 2020}{Seattle, WA, USA}
\acmBooktitle{Proceedings of the 28th ACM International Conference on Multimedia
	(MM '20), October 12--16, 2020, Seattle, WA, USA}
\acmPrice{15.00}
\acmDOI{10.1145/3394171.3413905}
\acmISBN{978-1-4503-7988-5/20/10}

\def\sexyname{KB-Ref\xspace}
\newcommand{\modelname}{ECIFA\xspace}
\newcommand{\ie}{\textit{i.e.}\xspace}
\newcommand{\eg}{\textit{e.g.}\xspace}
\newcommand{\etal}{\textit{et al.}\xspace}
\def\T{{\!\top}}
\begin{document}
	
	\fancyhead{}
	\title{Give Me Something to Eat: \\ Referring Expression Comprehension with Commonsense Knowledge}
	
	\author{Peng Wang}
	
	
	\affiliation{%
		\institution{Northwestern Polytechnical University}
	}
	
	\email{peng.wang@nwpu.edu.cn}
	
	%\author{Lars Th{\o}rv{\"a}ld}
	%%\affiliation{%
	%  \institution{The Th{\o}rv{\"a}ld Group}
	%  \streetaddress{1 Th{\o}rv{\"a}ld Circle}
	%\email{larst@affiliation.org}
	
	\author{Dongyang Liu}
	\affiliation{%
		\institution{Northwestern Polytechnical University}
	}
	\email{1226726279@mail.nwpu.edu.cn}
	
	\author{Hui Li}
	%\authornote{The corresponding author.}
	\affiliation{%
		\institution{The University of Adelaide}
	}
	\email{huili03855@gmail.com}
	
	\author{Qi Wu}
	\affiliation{%
		\institution{The University of Adelaide}
	}
	\email{qi.wu01@adelaide.edu.au}

	
	\begin{abstract}
		Conventional referring expression comprehension (REF) assumes people to query something from an image by describing its visual appearance and spatial location, but in practice, we often ask for an object by describing its affordance or other non-visual attributes, especially when we do not have a precise target. For example, sometimes we say 'Give me something to eat'. In this case, we need to use commonsense knowledge to identify the objects in the image. Unfortunately, there is no existing referring expression dataset reflecting this requirement, not to mention a model to tackle this challenge. In this paper, we collect a new referring expression dataset, called KB-Ref, containing $43$k expressions on $16$k images. In KB-Ref, to answer each expression (detect the target object referred by the expression), at least one piece of commonsense knowledge must be required. We then test state-of-the-art (SoTA) REF models on KB-Ref, finding that all of them present a large drop compared to their outstanding performance on general REF datasets. 
		We also present an expression conditioned image and fact attention (ECIFA) network that extracts information from correlated image regions and commonsense knowledge facts.
		Our method leads to a significant improvement over SoTA REF models,
		although there is still a gap between this strong baseline and human performance. The dataset and baseline models are available at: \textit{\url{https://github.com/wangpengnorman/KB-Ref\_dataset}}.
	\end{abstract}
	
	
	
	
	

	\maketitle
	
	\input{introduction.tex}
	
	\input{related-work.tex}

	\input{dataset.tex}

	\input{method.tex}

	\input{experiment.tex}

	\input{conclusion.tex}

	\begin{acks}
		Peng Wang, Dongyang Liu's participation in this work were in part supported by National Natural Science Foundation of China (NO.61876152, NO.U19B2037). Qi Wu is not supported by any of the projects above.
	\end{acks}
	
	\bibliographystyle{ACM-Reference-Format}
	\bibliography{sample-base}
	
	%\appendix
	
	
	
	%Lorem ipsum dolor sit amet, consectetur adipiscing elit. Morbi
	%malesuada, quam in pulvinar varius, metus nunc fermentum urna, id
	%sollicitudin purus odio sit amet enim. Aliquam ullamcorper eu ipsum
	%vel mollis. Curabitur quis dictum nisl. Phasellus vel semper risus, et
	%lacinia dolor. Integer ultricies commodo sem nec semper.
	
	
	%Etiam commodo feugiat nisl pulvinar pellentesque. Etiam auctor sodales
	%ligula, non varius nibh pulvinar semper. Suspendisse nec lectus non
	%ipsum convallis congue hendrerit vitae sapien. Donec at laoreet
	%eros. Vivamus non purus placerat, scelerisque diam eu, cursus
	%ante. Etiam aliquam tortor auctor efficitur mattis.
	
	
	%Nam id fermentum dui. Suspendisse sagittis tortor a nulla mollis, in
	%pulvinar ex pretium. Sed interdum orci quis metus euismod, et sagittis
	%enim maximus. Vestibulum gravida massa ut felis suscipit
	%congue. Quisque mattis elit a risus ultrices commodo venenatis eget
	%dui. Etiam sagittis eleifend elementum.
	
	%Nam interdum magna at lectus dignissim, ac dignissim lorem
	%rhoncus. Maecenas eu arcu ac neque placerat aliquam. Nunc pulvinar
	%massa et mattis lacinia.
	
\end{document}

%% file: introduction.tex
\begin{figure}[t]
	\setlength{\abovecaptionskip}{0.3cm}
	\centering
	\label{fig: example}
	\resizebox{0.37\textwidth}{!}{
		\includegraphics[width=\linewidth]{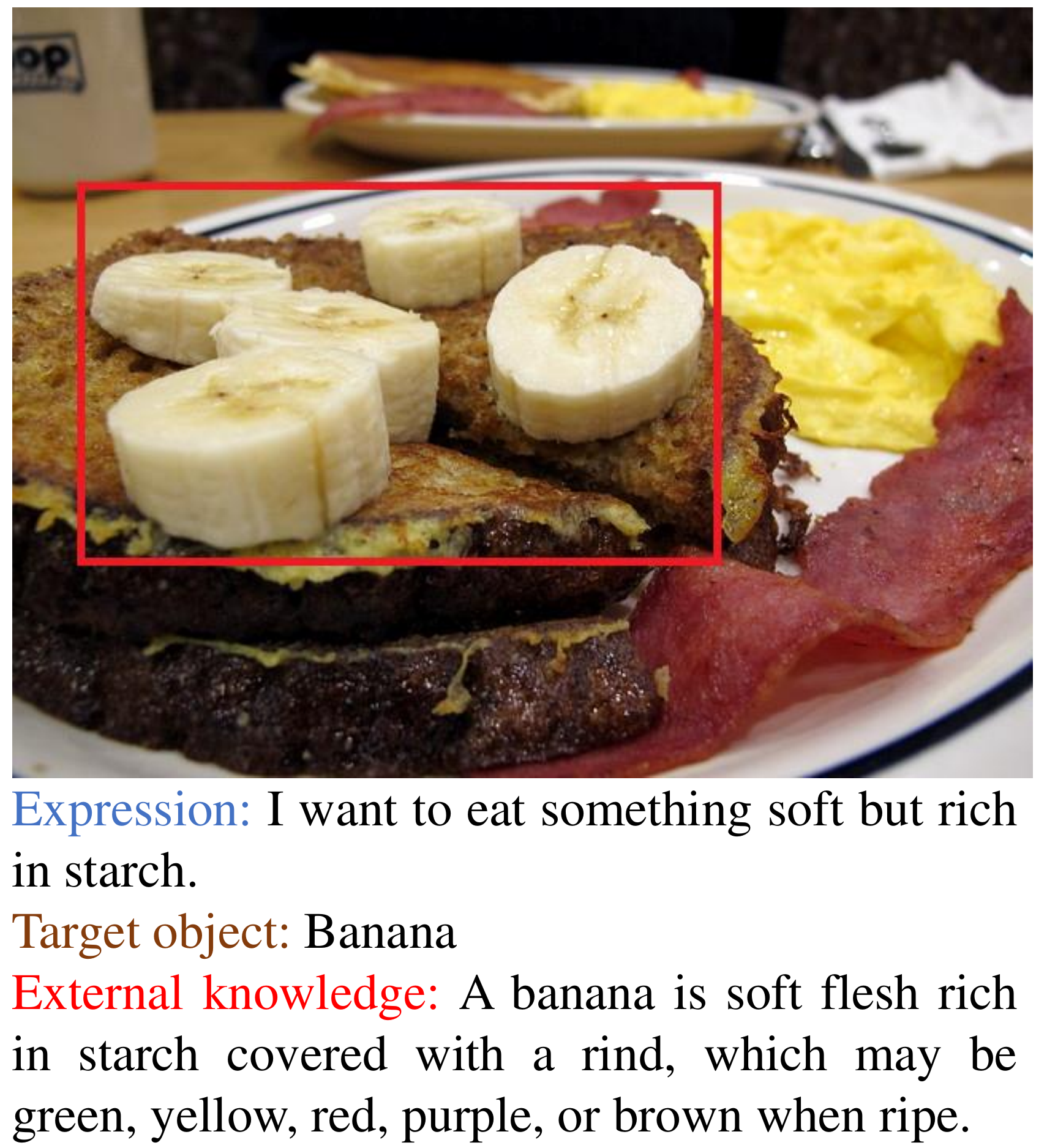}}
	\vspace{-4pt}
	\caption{An example from our KB-Ref dataset. 
		The key information in the expression, `soft but rich in starch', is non-visual attributes of the target object `banana', which can be 
		retrieved from an external knowledge base, such as ConceptNet.}
	\vspace{-1cm}
\end{figure}

% Introduce Knowledge into RE
% Dataset
% New Technical challenges
\section{Introduction}
% Referring expression and its issues
Referring expression comprehension (REF) aims at localizing a specific object in the image, based on an expression in the form of natural language. Several benchmark datasets have been released to test the referring expression comprehension models' ability, such as RefCOCO~\cite{kazemzadeh2014referitgame}, RefCOCOg~\cite{mao2016generation} and CLEVR-Ref+~\cite{liu2019clevr}.
Expressions in these existing datasets are usually about the visual appearance and spatial location of the target objects. For example, commonly seen expressions in RefCOCO~\cite{kazemzadeh2014referitgame} mainly include three components that are subject, location and relationship, where the subject component handles the visual categories, colour and other visual attributes; the location phrase handles both absolute and relative location; and the relationship covers subject-object visual relations~\cite{yu2018mattnet}, such as `the second white cup on the table'. Expressions in CLEVR-Ref+~\cite{liu2019clevr} require a longer reasoning chain but only visual attributes (such as size, colour, material) and spatial relationship (such as left, right) are covered.

While in practice, humans often use richer knowledge to ask for something they want, not limited to visual information. For example, we commonly use the `affordance' and other non-visual attributes to describe something we want, like `Can you pass me something to knock in this pin' and `I want to eat something low fat'. In this case, one needs to identify the objects in the image in accordance with the commonsense knowledge mentioned in the expression, for example a `rock' in the image can be used to knock in the pin, and `banana' is low fat. Thus, to enable a machine to reason over visual appearance, spatial and semantic relationships and commonsense knowledge is an emerging challenge.

Unfortunately, no existing datasets, including the popular RefCOCOg \cite{mao2016generation} and CLEVR-Ref+~\cite{liu2019clevr}, present above features, not to mention a referring expression model that offers this capacity. 
To this end, we propose a new dataset for referring expression comprehension with commonsense knowledge, {\bf KB-Ref}, collected based on the images from Visual Genome \cite{krishna2017visual} and knowledge facts from Wikipedia, ConceptNet~\cite{speer2017conceptnet} and WebChild~\cite{Tandon2017WebChild}. Similar to RefCOCO dataset family, we ask Amazon Mechanical Turk (MTurk) workers to select an object from the image and use language to describe it so that another person can use it to localize the object. The difference in annotation is that we also provide a list of commonsense facts about the selected object to workers, who must use at least one of the provided knowledge facts together with the visual context to describe the target object. We also ask workers to record the knowledge they used. To verify whether the collected expression is meaningful and whether the recorded knowledge is required to localize the object, we give the annotations to another MTurk group to verify. Only those expressions that need knowledge to solve are kept. This leads to $43,284$ expressions of $1,805$ object categories in $16,917$ images. The average length of the expressions is $13.32$, nearly double the length in RefCOCOg~\cite{mao2016generation}. In our setting, the recorded knowledge facts are provided during the training but are removed during the testing. So the real technical challenge of this new task is how to mine related knowledge and combine it with visual context to find the object that is referred by the expression.

To verify whether commonsense knowledge is crucial in our collected dataset, we first evaluate a variety of state-of-the-art (SoTA) referring expression models (such as MattNet~\cite{yu2018mattnet} and LGARNs~\cite{wang2019neighbourhood}) on our KB-Ref dataset, finding that all of them present a large drop compared to their performance on general REF datasets. 
We then propose an {\bf E}xpression {\bf C}onditioned {\bf I}mage and {\bf F}act {\bf A}ttention network ({\bf \modelname}), 
which uses an top-down attention module to extract expression-related image representations 
and an episodic memory module to focus attention on a subset of commonsense knowledge facts.
The proposed network leads to a significant improvement over SoTA REF methods, on our constructed dataset.
Nevertheless,
we also evaluate the human performance on the test split and find that there is still a large gap between our baselines and the human accuracy, which suggests that our proposed KB-Ref dataset is considerably challenging.
%CNN-RNN baseline (termed as \modelname) with a facts memory network that uses the expression to retrieve related knowledge for identifying the object. This module can be plugged into both our baseline model and SoTA REF models, 
%leading to a significant and consistent improvement. 

% List of contributions

%% file: related-work.tex
\section{Related Work}
\subsection{Referring Expression Comprehension}
\paragraph{Datasets.}

\begin{table*}
	\newcommand{\tabincell}[2]{\begin{tabular}{@{}#1@{}}#2\end{tabular}}
	\begin{center}
		%\addtocounter{footnote}{1} 
		%\footnotetext{\footnotemark{footnote}}  
		\scalebox{0.8}{
			\begin{tabular}{|l|c|c|c|c|c|c|c|}
				\hline
				Dataset & Facts & Images & {\tabincell{c}{Objects \\ Categories}}  & {\tabincell{c}{Referring \\ Expressions}} & {\tabincell{c}{Objects \\ Instances}} & {\tabincell{c}{Objects Queried \\ Per Image}} & {\tabincell{c}{Average \\ Expression\\ length}}\\
				\hline\hline
				RefCOCO \cite{kazemzadeh2014referitgame} & $\times$ &  $19,944$ & $78$ & $142,210$ & $50,000$ & $2.51$ & $10.19$ \\
				RefCOCO+ \cite{kazemzadeh2014referitgame} & $\times$ & $19,992$ & $78$ & $141,564$  & $49,856$ & $2.49$ & $10.14$ \\
				RefCOCOg \cite{mao2016generation}$^\dag$ & $\times$ & $25,799$ & $78$ & $95,010$  & $49,822$ & $1.93$ & $8.93$ \\
				%CLEVR-Ref+ \cite{liu2019clevr} & $\times$ & $85,000$ & $8$ & $848,535$  & $848,535$ & $9.98$ & $28.66$ \\
				{KB-Ref} (ours) & $\checkmark$ &  $16,917$ & $1,805$$^\ddag$ & $43,284$  & $43,284$ & $2.56$ & $13.32$ \\
				\hline
			\end{tabular}}
		\end{center}
		\caption{Comparison between different REF datasets. $^\dag$The data of RefCOCOg is based on the revised version. 
			$^\ddag$We use Visual Genome's definition of object categories, which is finer than COCO. For example, `people' in COCO corresponds to `man', `woman', `boy' and `girl' in Visual Genome.}
		\label{datasetsComp}
		\vspace{-0.8cm}
	\end{table*}

	As presented in Table~\ref{datasetsComp}, commonly used datasets for referring expression comprehension include RefCOCO~\cite{kazemzadeh2014referitgame}, RefCOCO+~\cite{kazemzadeh2014referitgame} and RefCOCOg~\cite{mao2016generation}, all collected on top of MSCOCO images~\cite{MSCOCO}. RefCOCO and RefCOCO+ are collected interactively in a two-player game, with concise phrase descriptions, while RefCOCOg is collected by MTurk workers in a non-interactive setting, using longer declarative sentences. 
	There is no restriction in RefCOCO on language expressions, while RefCOCO+ focuses more on purely appearance descriptions where location words are not allowed.  
	GuessWhat?!~\cite{GuessWhat} is another dataset based on MS-COCO images. Instead of using a single expression, it creates a sequence of sentences (\ie, dialog) for a given image to perform referring expression comprehension. 
	CLEVR-Ref+~\cite{liu2019clevr} is a recently introduced synthetic dataset, built on the CLEVR environment. 
	%It is a synthetic diagnostic dataset , which involves very complicated queries aiming at assess the relational reasoning ability of REF models. 
	{ In contrast, our proposed KB-REF is based on images from Visual Genome~\cite{krishna2017visual}, which provides richer annotations including objects, attributes and relationships. The expressions in KB-REF are different with above and need both visual context and commonsense knowledge to resolve.}
	
	\paragraph{Approaches.}
	Referring expression is a visual-linguistic cross-model understanding problem. Some works solve REF jointly with a referring expression generation task~\cite{kazemzadeh2014referitgame, mao2016generation, GroundeR, yu2016modeling}.
	%, and look for the object region that can produce the most similar description to the given expression. 
	Some others~\cite{hu2017modeling,Luo2017,niu2019variational,wang2019neighbourhood,yu2018mattnet} propose different types of joint embedding frameworks and directly localize the object which has the highest matching score. 
	%Some approaches use single feature vectors to represent expressions and image regions, which however ignore the structural relations in language and image. 
	%To this end, 
	The work in~\cite{hu2017modeling,yu2018mattnet} proposes to decompose the expression into sub-phrases, which are then used to trigger separate visual modules to compute matching score. Liu~\etal~\cite{DaqingICCV2019} develop a neural module tree network to regularize the visual grounding along the dependency parsing tree of the sentence. The works in~\cite{Chaoruicvpr2018,SibeiICCV2019,Bohancvpr2018} argue to learn the representations from expression and image regions in a stepwise manner, and perform multi-step reasoning for better matching performance.
	%, using different types of attention mechanisms. 
	Wang~\etal~\cite{wang2019neighbourhood} propose a graph-based language-guided attention network to highlight the inter-object and intra-object relationships that are closely relevant to the expression for better performance. Niu~\etal~\cite{niu2019variational} develop a variational Bayesian framework to exploit the reciprocity between the referent and context. { Our proposed baseline model uses a simple visual-expression joint embedding model to calculate the matching score, but we incorporate a knowledge facts attention module to mine external knowledge base for referring expression, which can also be plugged into other REF models for extra knowledge exploring.}
	
	\subsection{Visual Understanding and Reasoning with Extra Commonsense Knowledge}
	%The use of external knowledge has led to a great development in NLP domain such as semantic parsing~\cite{}, information retrieval~\cite{}, \etc.
	
	%TODO: Add more Knowledge-Required Visual Reasoning works
	Commonsense knowledge has already attracted research attention in visual understanding and reasoning, such as visual relationship detection~\cite{lu2016visual, yu2017visual, xu2019learning}, 
	scene graph generation~\cite{gu2019scene}, visual question answering~\cite{askmeanything, liu2019clevr, Xiong2016Dynamic, FVQA, narasimhan2018straight, Narasimhan2018, KDMN, hudson2019gqa} and zero-shot recognition~\cite{wang2018zero, lee2018multi}.
	In particular, the incorporation of commonsense knowledge is important for
	visual question answering (VQA), because a lot of questions are from open-domain that require to perform reasoning beyond the image contents. In~\cite{askmeanything}, attributes extracted from the image are used to query external knowledge based on DBpedia, which enables answer questions beyond the image. Dynamic memory networks are employed in~\cite{VKMN} and~\cite{KDMN} to incorporate structured human knowledge and deep visual features for answer decoding. 
	Wang~\etal~\cite{FVQA} introduce a Fact-Based VQA (FVQA) dataset which requires external knowledge to answer a question. A Graph Convolution Network (GCN) is developed in~\cite{Narasimhan2018} which integrates image, question and all possible facts in an entity graph for answer inferring in FVQA. The work in~\cite{licvpr2019} represents visual information by dense captions and convert VQA as a reading comprehension problem, where extra knowledge is added by text concatenation. In comparison with FVQA where visual reasoning is based on fixed structured knowledge bases, an even larger knowledge based VQA dataset (OK-VQA) is introduced in~\cite{OKVQA} which performs VQA over unstructured open knowledge. 
	The Visual Commonsense Reasoning (VCR)~\cite{zellers2019recognition, hudson2019gqa} dataset contains
	290k multiple choice QA problems from 110k movie scenes, which require
	higher-order cognition and commonsense reasoning.
	
	Similarly, as a proxy to evaluate AI systems on both vision and language understanding, REF would require not only object localization based on image appearance, but also a more natural way to achieve human-level semantic understanding. With this requirement, we go one step forward and design a KB-Ref which requires reasoning incorporating external commonsense knowledge. In particular, REF can be regarded as a subtask of VQA with the question as ``\textbf{Where} is \textit{sth. (by referring expression)} in the image?''. However, in VQA the answer is generally open-ended, presented in natural language, while in REF the answer is numerical, chosen from a group of candidate bounding boxes, or directly output a detected one, which makes the evaluation much easier.

%% file: dataset.tex
\begin{figure}[htbp]
	\setlength{\abovecaptionskip}{0.3cm}
	\centering
	\includegraphics[width=0.47\textwidth]{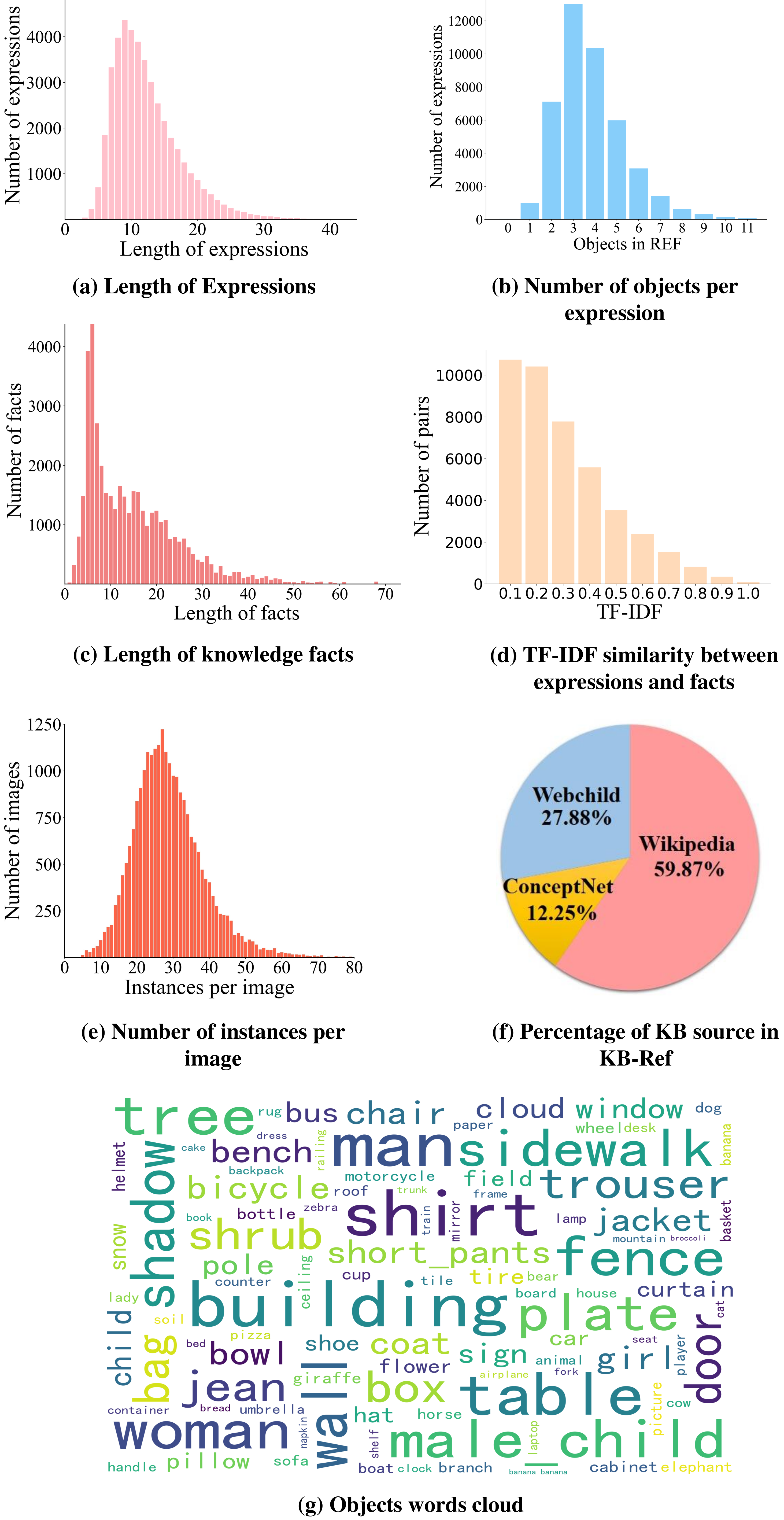}
	\caption{Statistical analysis of the proposed \sexyname dataset. We show the distributions of the length of expressions, the number of objects mentioned per expression in and the the length of knowledge facts in subfigures (a), (b) and (c) respectively. 
	We also collect statistics of the TF-IDF similarity (the smaller the number, the less similar) between expressions and their corresponding supporting facts in subfigure (d) to see how much the expressions differ from facts. The number of object instances per image are also counted whose statistical result is shown in subfigure (e). The percentages of 
	where the groundtruth facts come from (KB source) are illustrated in subfigure (f). 
	The word cloud of the queried object categories is shown in subfigure (g), where the font size indicates the corresponding number of expressions.}
	\label{statistic}
\end{figure}

\section{The KB-Ref Dataset}

Different from existing referring expression datasets~\cite{kazemzadeh2014referitgame, liu2019clevr, mao2016generation, yu2016modeling} that mainly cover the visual contents of the referred objects (such as appearance, attributes and relationships), we collect a new dataset called KB-Ref that needs additional commonsense knowledge to identify the referent object. Wikipedia, ConceptNet and WebChild are employed here as knowledge resources. In this section, we will describe our data collection pipeline in detail and give a statistic analysis of the dataset.

\subsection{Dataset Collection}
\paragraph{Images and Objects} Images of our dataset are sampled from Visual Genome (VG) \cite{krishna2017visual}, which contains over $108$K images with dense annotations on objects, attributes and relationships. There are averagely $36.5$ bounding boxes in each image, which requires complex visual reasoning to localize.
Descriptions on object's affordance or other non-visual attributes can help the referring expression comprehension, which, however, usually needs commonsense knowledge to understand. In VG, most objects are canonicalised to a synset ID in WordNet~\cite{miller1995wordnet}. In order to complicate our dataset, we ignore objects that do not belong to any synset. Same as RefCOCO \cite{kazemzadeh2014referitgame} and other REF datasets, objects that appear alone in one image (\ie, there are no other instances of the same object category within the same image) are also removed. Moreover, we neglect objects whose shorter sides are less than $32$ pixels. Then we eliminate images which has labeled objects less than $5$. With those filtering process, there are $24,453$ images left which have $2,075$ object categories within $208,532$ bounding boxes. They form the basis of our KB-REF dataset.

\paragraph{Knowledge Base} 
In order to aid annotation and evaluation, we construct a knowledge base by collecting facts from three knowledge resources (\ie, Wikipedia, ConceptNet and WebChild) that are related to the $1,805$ object categories appeared in our dataset. 
ConceptNet is a graph-structured commonsense knowledge base where facts are represented by triplets of start nodes, relations, and end nodes. There is a closed set of relations in ConceptNet, such as \texttt{IsA}, \texttt{HasA}, \texttt{PartOf}, \texttt{MadeOf}, \texttt{UsedFor}, and \texttt{CapableOf}.
WebChild contains fine-grained commonsense knowledge distilled from web-scale amounts of text, in which the facts can be further categorized into properties (\eg, \texttt{HasShape}, \texttt{HasSzie}, \texttt{HasTaste}), comparative (\eg, \texttt{FasterThan}, \texttt{SmallerThan}) and part-whole (\eg, \texttt{PhysicalPartOf}, \texttt{SubstanceOf}, \texttt{MemberOf}).
Compared to structured knowledge bases like ConceptNet and WebChild,
Wikipedia contains a larger variety of knowledge but in unstructured format.
For each object category, we collect the facts in ConceptNet and Webchild whose start nodes or end nodes match the category label, and the Wikipedia article
whose theme concept corresponds to this category
\footnote{Considering that some object categories have different interpretations in different images, \eg, a pole may be a ski pole or a bar holding something, in this case, we choose the meaning that appears most frequently in our dataset, and ignore the uncommon ones. But this issue is then fixed in the following human annotation section because we allow human workers to fix or even rewrite the required knowledge.}.
To unify the fact format, we translate the triplet fact in ConceptNet and Webchild into sentences, and treat each sentence in Wikipedia articles as a fact.

\paragraph{Data Annotation} We ask Amazon Mechanical Turk (MTurk) workers to write down referring expressions for the queried objects.  
The following requests are put forward in the annotation process. {1)} At least one fact from the constructed knowledge base should be used in referring expression. {2)} The specific object name cannot appear in the expression. Annotators are required to describe the queried object based on the corresponding fact and its visual context. {3)} Multiple auxiliary objects appeared in the image are encouraged to be mentioned in the expression, to aid the search of the target object.

To control the dataset bias, we also perform a quality check in background: 

The frequency of each fact adopted in the expressions cannot exceed $200$. 
If exceeds, this fact will be removed in the following annotation process. 
Note that we use TF-IDF (Term Frequency–Inverse Document Frequency) to measure the similarity between facts. If TF-IDF between two facts is larger than $0.5$, they are regarded as the same.

The detailed annotating process is as follows. Given the object to be queried in an image (highlighted by a bounding box) and the related facts in our knowledge base, the MTurk worker is asked to generate a unique text description about the object, according to the requests described above. It takes about $2$ minutes to collect one expression. Then another annotator is asked to verify the correctness of the provided expression. The ones that do not conform to the requests will be asked to re-annotate. The annotation interface can be found in the supplementary materials.

\subsection{Data Analysis}
Totally, we collected $43,284$ expressions for $1,805$ object categories on $16,917$ images, as compared with other datasets listed in Table~\ref{datasetsComp}.  Each object instance in one image has a sole referring expression. To be specific, Figure~\ref{statistic}(a) shows the distribution of expression lengths. The average length of referring expression in \sexyname is $13.32$ words, which is longer than that in RefCOCOs (including RefCOCO, RefCOCO+ and RefCOCOg) (about $10$).
%, but shorter than that in CLEVER-Ref+ ($28.66$), which mainly focuses on complex visual  relationship reasoning. 
In our dataset, $25,626$, $9,045$ and $8,613$ expressions are generated based on the knowledge facts from Wikipedia, ConceptNet and WebChild, respectively.
Figure~\ref{statistic}(b) shows the distribution of the number of objects mentioned in each expression. Averagely, there are $4.34$ objects used per expression, which suggests the complexity of our collected expressions. 
The distribution of the length of fact sentences is presented in Figure~\ref{statistic}(c), with an average of $16.78$ words per fact, which reflects the rich information recorded in these facts. 
%suggests that the unstructured commonsense knowledge is not easy to be understood. 
We also use TF-IDF to calculate the similarity between each expression and the corresponding fact. As shown in Figure~\ref{statistic}(d), most TF-IDFs range from $0.1$ to $0.4$, which illustrates the difference between the expressions and their corresponding facts. 
Note that our collected expressions not only reflect the knowledge from their corresponding facts but also contain visual information about the target objects. 

Figure~\ref{statistic}(e) shows the number of instances per image. We can see most of the images include multiple objects ranging from $20$ to $50$. The Figure~\ref{statistic}(f) shows the percentage of knowledge sources of our dataset, most of the facts are from wikipedia. The object category cloud shown in Figure~\ref{statistic}(g) illustrates that our dataset covers a wide range of objects with less bias (the font size in the cloud represents the frequency of the object appeared in our dataset).

We split the dataset on the base of images randomly for training, validation and test. 
There are $31,284$ expressions with $9,925$ images in training set, $4,000$ expressions with $2,290$ images in validation set, and $8,000$ expressions with $4,702$ images in test set.

%% file: method.tex
\begin{figure*}
	\centering
	\scalebox{0.9}{
		\includegraphics[width=\linewidth]{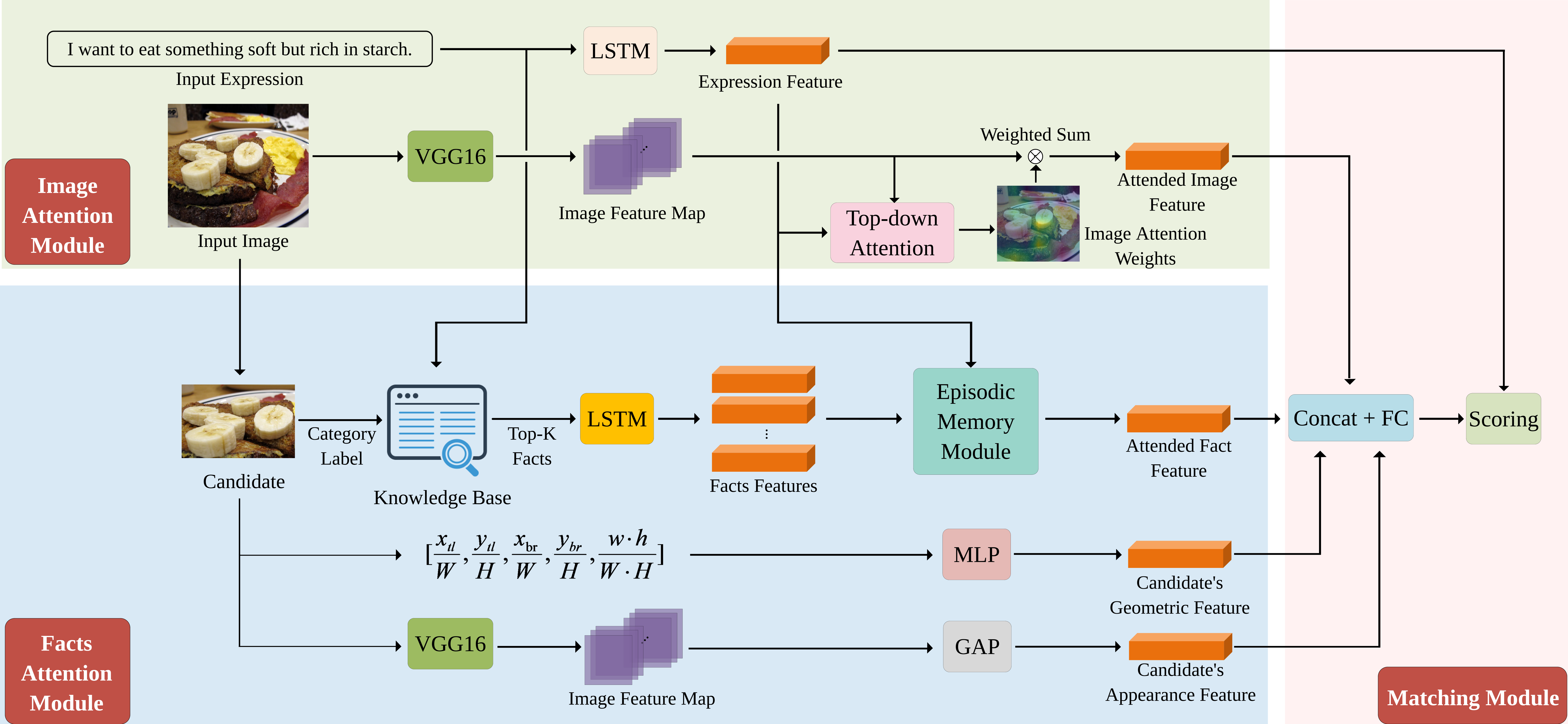}
	}
	\vspace{-4pt}
	\caption{The overall architecture of our baseline model, which contains three main parts, \ie, the top-down attention module, the facts attention module and the matching module. These modules will be described one by one in Section~\ref{method}.
	}
	\label{fig:arch}
	\vspace{-0.5cm}
\end{figure*}

\section{Method}
\label{method}

%In this section, we introduce our knowledge dynamic memory network (\modelname) model for extra knowledge required referring expression comprehension. 
In this section, we propose an Expression Conditioned Image and Fact Attention (\modelname) network for extra knowledge required referring expression comprehension. 
Given a natural language expression $q$ and an image $I$, the model is asked to pick the described object $O^*$ from a group of candidates $\{O\}_{n=1}^N$. The bounding boxes of candidate objects are either groundtruth or obtained via off-the-shell detectors. Different from previous settings, extra commonsense knowledge is needed to understand the given expression for object grounding.
The overall architecture is illustrated in Figure~\ref{fig:arch}. %Our model is inspired by GroundeR~\cite{GroundeR}, a simple baseline model in referring expression, 
%but it is tailored to incorporate a group of facts extracted from an extra knowledge base. 
The model can be generally divided into three components: 
(1) a top-down image attention module that predicts an attention distribution over the  image grids conditioned on the given expression; 
(2) a multi-hop facts attention module that gather information from a set of related facts in our knowledge base; 
(3) a matching module which calculates the expression-object matching score for final grounding. We elaborate on each component in the following. It is worth noting that our facts attention module can be plugged into other referring expression models as well.

\subsection{Top-down Image Attention Module}
Similar to many conventional REF models~\cite{Hu2016cvpr, mao2016generation}, we first represent each word in the given expression $q$ using an one-hot vector, and then encode them iteratively by an LSTM. The hidden states at all time steps 
%(denoted as $h_t^q$, $t=1,\dots,T$, where $T$ is the length of the expression) 
are added together\footnote{We also tried to use the last hidden state of the LSTM as the expression feature but the results are slightly worse. We believe the reason is that our expressions are long.} as the holistic representation for the expression, which is denoted as $\mathbf{q}$ with a dimension of $2048$. Meanwhile, the input image is fed into a pre-trained VGG-$16$ net. Feature maps from Conv${5\_}3$ are extracted, denoted as $\mathbf{V}$ of size $7\times7\times512$. 
A top-down attention mechanism is adopted here to extract information 
from the image regions that are the most related to the expression, which is formulated as:
\begin{flalign}\label{globalatt}
	\begin{split}
		{\alpha}_{i,j} &= \mathbf{w}^\T \tanh(\mathbf{W}_v \mathbf{V}_{i,j} + \mathbf{W}_q \mathbf{q}), \\
		\beta_{i,j} &= \exp(\alpha_{i,j}) / (\textstyle{\sum_{k,l}} \exp(\alpha_{k,l})),  \, i,j \in \{1,...,7\}\\
		\mathbf{v} &= \textstyle{\sum_{k,l}} \beta_{k,l} \mathbf{V}_{k,l},
	\end{split}
\end{flalign}
where $\mathbf{V}_{ij}$ is the local feature vector at position $(i,j)$ in feature maps $\mathbf{V}$; The expression feature $\mathbf{q}$ is used here as the guidance signal; $\mathbf{W}_v$, $\mathbf{W}_q$, and $\mathbf{w}$ are linear transformation weights to be learned; $\beta_{ij}$ is the attention weight at location $(i,j)$. The weighted sum $\mathbf{v}$ is the attended image feature, with the dimension of $512$. It encodes image features that is most relevant to the given expression.

\subsection{Two-stage Fact Attention Module}
\label{sec:dmn}
The distinguishing feature of our proposed dataset is the requirement of commonsense knowledge.
In this section, we introduce a two-stage coarse-to-fine fact attention module that distills
related information from the massive facts of our constructed knowledge base.

For the first stage, we train a Word2Vec~\cite{Mikolov2013Distributed} model with Skip-Gram on the $1,008,406$ facts in our knowledge base. 
Given a candidate object, we first retrieve its corresponding facts, and then compute the cosine similarity between the averaged Word2Vec word embeddings of each fact and the expression.
At most top $K$ facts (denoted as $\{s_k\}, k=1, \dots, K$) are then kept for further processing.

\begin{figure}
	\centering
	\includegraphics[width=0.95\linewidth]{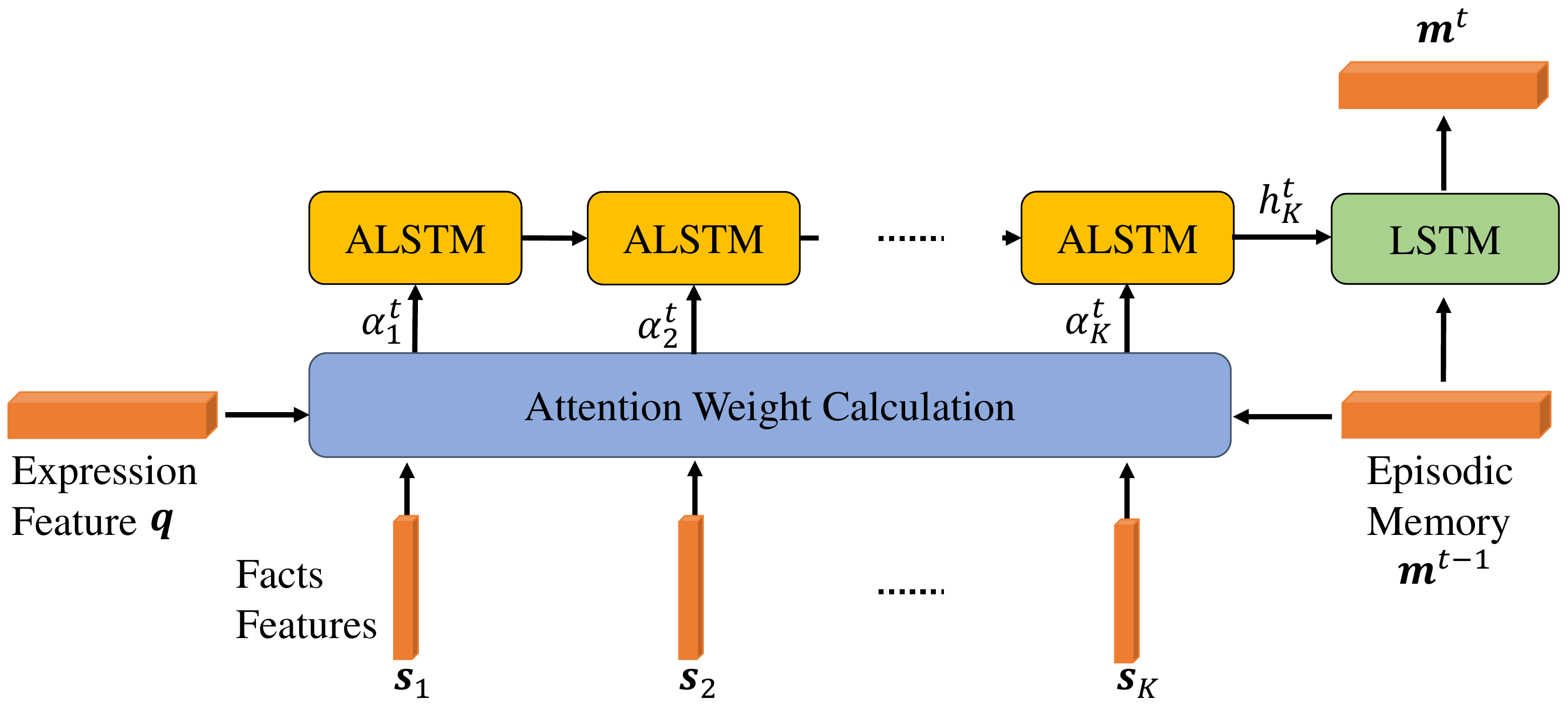}
	\vspace{-1em}
	\caption{The computational flow chart of the episodic memory module at pass $t$.}
	\label{fig:mn}
	\vspace{-0.5cm}
\end{figure}

At the second stage, inspired by \cite{Xiong2016Dynamic}, we employ an Episodic Memory Module (as shown in Figure \ref{fig:mn}) to focus attention on a subset of the $K$ retrieved facts in the previous stage.
Firstly, each fact $s_k$ is encoded by an LSTM with $2048$D hidden states (which does not share parameters with the LSTM encoding expressions), and the averaged hidden states over all time steps (denoted as $\mathbf{s}_k$) is taken as the fact representation, considering that some facts are very long.  
%The encoded fact features for object $O_m$ are denoted as $\mathbf{S}_m \in \mathcal{R}^{2048 \times K}$, where each column vector $\mathbf{s}_{m,k}$ corresponds to the $k$th fact. 
Next, the episodic memory module is adopted to perform a multi-hop attention over facts
$\mathbf{s}_1, \dots, \mathbf{s}_K$ under the guidance of the expression $\mathbf{q}$.
At each pass $t$, a set of attention weights are computed as follows:
\begin{flalign}\label{mnm_1}
	\begin{split}
		\mathbf{z}_{k}^t &=  [\mathbf{s}_{k} \circ \mathbf{q};\mathbf{s}_{k} \circ \mathbf{m}^{t-1}; |\mathbf{s}_{k} - \mathbf{q}|;|\mathbf{s}_{k} - \mathbf{m}^{t-1}|], \\
		z_{k}^t &= \mathbf{w}_{z}^{\T} \tanh(\mathbf{W}_{z} \mathbf{z}_{i, k}), \\
		{\alpha}_k^t &= \exp({z_k^t}) / (\textstyle{\sum}_{l=1}^K \exp({z_l^t})), \, k = 1, \dots, K,
	\end{split}
\end{flalign}
which is then fed into an attentional LSTM to decide how much the hidden state should be updated for each $k$:
\begin{flalign}\label{mnm_2}
	\begin{split}
		%\mathbf{h}_{i,k} &= RNN(\mathbf{s}_{k}, \mathbf{h}_{i,k-1})\\
		\mathbf{h}_{k}^t &= \alpha_{k}^t \mathrm{LSTM}(\mathbf{s}_{k}, \mathbf{h}_{k-1}^t) 
		+ (1-\alpha_{k}^t) \mathbf{h}_{k-1}^t, \, k = 1, \dots, K\\
		% \mathbf{m}_{i+1} &= LSTMCell(\mathbf{e}_{i,K}, (\mathbf{m}_{i}, 0))\\
	\end{split}
\end{flalign}
and the episodic memory for pass $t$ is updated by another LSTM that takes the last hidden state of the attentional LSTM as contextual vector:
\begin{flalign}\label{mnm_3}
	\begin{split}
		\mathbf{m}^{t} &= \mathrm{LSTM}(\mathbf{h}_{K}^t, \mathbf{m}^{t-1}), \, t = 1, \dots, T.\\
	\end{split}
\end{flalign}
The memory for the last pass $\mathbf{m}^{T}$ is considered as the attended fact feature and fed into the following Matching Module.  

\subsection{Matching Module}
The matching module is then used to calculate the matching score between the expression $q$ and each object $O_n$. Specifically, for each candidate object, we calculate its appearance feature by firstly resizing the object region to $224 \times 224$ and then feeding it into a pre-trained VGG-$16$. Feature maps from Conv${5\_3}$ are extracted and averagely pooled. A fully connected layer with $512$ neurons and ReLU are then followed, which results in an appearance feature for object $O_n$ of $\mathbf{f}_n^a \in \mathbb{R}^{512}$. 

In addition, we also extract the geometric information for each candidate object, $[\frac{x_{tl}}{W}, \frac{y_{tl}}{H}, \frac{x_{br}}{W}, \frac{y_{br}}{H}, \frac{w\cdot h}{W\cdot H}]$, which is a $5$-dimensional vector consisting of four values for top left and bottom right corner coordinates of the object region (normalised between
0 and 1) and one value for its relative area (\ie, ratio of the bounding box area to the image area, also between 0 and 1). A fully connected layer with $128$ neurons and ReLU are followed, which lead to a geometric feature $\mathbf{f}_n^p \in \mathbb{R}^{128}$.

The candidate's appearance feature $\mathbf{f}_n^a$ and geometric feature of $\mathbf{f}_n^g$ are then concatenated with the attended fact feature $\mathbf{f}_n^e$ and the attended image feature $\mathbf{v}$. Another linear transformation is applied to yield a $2048$d feature: 
\begin{equation}
	\mathbf{f}_n = 
	\mathbf{W} [\mathbf{f}_n^a; \mathbf{f}_n^g; \mathbf{f}_n^e; \mathbf{v}],
	\label{eq:fmall}
\end{equation}
where $\mathbf{W}$ is the parameter to be learned. 
Finally we calculate the inner product of 
the expression feature $\mathbf{q}$ and 
the integrated object feature $\mathbf{f}_n$. Softmax is then applied over all candidate objects, 
and the object with the highest score will be selected for the expression.
During the training, the cross entropy loss is used.

%% file: experiment.tex
\section{Experiment}

In this section, we conduct experiments to analyze the proposed dataset \sexyname and baseline model \modelname. 
%The implementation details are introduced firstly. 
Firstly, we analyze the bias of our dataset by evaluating our algorithm with different partial input information. Then the proposed \modelname is compared with SoTA REF models on our dataset. Lastly, a group of ablation experiments are performed to validate the effectiveness of multi-hop fact attention. Additionally, results of using detected bounding boxes are given.

%a group of ablation experiments are performed to validate the effectiveness of multi-hop fact attention. 
%Then the proposed \modelname is compared with SoTA REF models on our dataset, using both groundtruth and detected boxes.
%Lastly, we analyze the bias of our dataset by evaluating our algorithm with different partial input information.

%Then, a group of ablation experiments are performed to analyze the dataset bias and the effectiveness of the proposed \modelname. Lastly we test \modelname~ as well as some SoTA REF methods on the proposed KB-Ref with ground truth bounding box or the bounding box that is produced by a detector.

%\subsection{Implementation Details}
%\label{impDetail}
All the experiments are conducted on $8$ Nvidia RTX2080Ti GPUs. The baseline model is implemented with PyTorch, and trained by using SGD optimizer with a learning rate of $1e^{-4}$ initially. The learning rate will decay half if the validation loss does not decrease in consecutive two epochs. we adopt a batch size of $16$, which consists of $16$ expressions and the corresponding object candidates in images, and train the model with $40$ epochs. Same as previous work, we also use accuracy as the evaluation metric, which is calculated by checking whether the target object is correctly selected or not.

\begin{table}
	{
		\begin{center}
			\scalebox{0.8}{
				\begin{tabular}{l|c|c|c|c}
					\hline
					Method & \multicolumn{2}{|c}{Accuracy (\%)} & \multicolumn{2}{|c}{FG Accuracy (\%)}  
					\\ \cline{2-5}	&  Val & Test &  Val & Test\\
					\hline
					Random & $9.93$ & $9.81$  & - & -\\
					\modelname~ (no image) & $49.73$ & $47.61$  & $41.91$ &  $40.35$\\
					\modelname~ (no facts) & $37.95$ & $35.16$  & - & - \\
					%\modelname~ (shuffle) & $-$ & $-$   & $-$ &  $-$\\
					\modelname~ (partial expression)& $59.07$ & $58.49$ &  $48.97$ & $48.61$\\
					\modelname~ & $59.45$ & $58.97$ &  $49.26$ &  $48.92$\\
					\hline
				\end{tabular}
			}
		\end{center}
	}
	\caption{Dataset bias analysis with different settings. The results show that visual and knowledge facts are both important to our dataset. 
		%The perturbations on referring expressions have relatively small influence on model performance. 
		The remove of prepositional phrases and verbs from expressions has a relatively small influence on model performance. 
		FG: fact grounding.}
	\label{table:bias}
	\vspace{-0.9cm}
\end{table}

\subsection{Dataset Bias Analysis}
Dataset bias is an important issue of current vision-and-language datasets.
In~\cite{cirik2018visual}, Cirik~\etal shows that a system trained and tested on input images without the input referring expression can achieve an accuracy of $71.2 \%$ in top-2 predictions on RefCOCOg~\cite{mao2016generation}, which suggests the significant data bias.
Inspired by this work, we analyze our dataset using similar methods. 

\noindent\textbf{Random} $\,$ The accuracy is obtained by selecting a random object from the candidates in an image. 

\noindent\textbf{\modelname~ (no image)} $\,$ We eliminate all the visual features in \modelname, \ie, $\mathbf{f}_n^a$, $\mathbf{f}_n^g$ and $\mathbf{v}$ are removed from Equation~\ref{eq:fmall} when calculating the matching score. This study is to investigate the importance of visual information in our Kb-Ref.

\noindent\textbf{\modelname~ (no facts)} $\,$ \modelname is re-trained without using knowledge facts, which means that $\mathbf{f}_m^e$ is removed from Equation~\ref{eq:fmall}, so as to study the impact of knowledge facts in our Kb-Ref.

%\noindent\textbf{\modelname~ (shuffle)} $\,$ We re-train \modelname with the words in referring expression shuffled as input. This experiment analyzes the influence of language syntactic structure in our Kb-Ref.

\noindent\textbf{\modelname~ (partial expression)} $\,$ \modelname is re-trained by keeping only nouns and adjectives in the input expression, since description words (\eg, color, shape) and object categories are basically expressed by adjectives and nouns. It will obscure the relationships between objects, which are usually represented by prepositional phrases and verbs.

Table~\ref{table:bias} shows the ablation study results. The \textbf{Random} baseline offers accuracy around $10\%$ on both validation and test sets, as there are around $10$ candidates to be selected for each expression. Without using external knowledge, the accuracy of \textbf{\modelname~ (no facts)} drops to $35.16 \%$ on test set. %, which is even lower than those produced by SoTA methods listed in Table~\ref{table:accuracy}. 
\textbf{\modelname~ (no image)} leads to a $11$-percentage drop on the test accuracy.
As we can see, the performance drop caused by removing facts is larger than by removing image, which indicates the importance of commonsense knowledge in our REF setting. %Compared to full \modelname, \textbf{\modelname~ (shuffle)} has small performance improvement. 
%We infer that the model does not rely much on the syntactic structure of the expression. The holistic representation $\mathbf{f}^q$ may only encode some shallower correlations between words. 
In addition, by discarding all words except nouns or adjectives, the test accuracy of \textbf{\modelname~ (partial expression)}  drops slightly by around $0.5$ percentage. %which is even lower than that obtained by \textbf{\modelname~ (shuffle)}. 
%It is still higher that the compared SoTA methods in Table~\ref{table:accuracy}.
This phenomenon indicates that the superior performance of \modelname~ does not specifically depend on object relationships.

Besides the answering accuracy, we also evaluate the accuracy of `fact grounding' in Table~\ref{table:bias}. A success will be counted if our model gives the highest attention weight to the groundtruth fact. We can see a positive correlation between the answering accuracy and the `fact grounding' accuracy.

\subsection{Comparison with State-of-the-art}
The following models are evaluated on the KB-Ref and compared with the proposed \modelname model.
All the models are trained from scratch on the training split of our proposed \sexyname dataset, using their own training strategies.

\noindent\textbf{CMN} \cite{hu2017modeling} is a modular architecture that utilizes the language attention to parse the input expression into subject, relation and object. The textual components are then aligned with image regions by three modules respectively to calculate the final matching score.

\noindent\textbf{SLR} \cite{yu2017joint} is a speaker-listener model that jointly learns for referring expression comprehension and generation. A reinforce module is introduced to guide sampling of more discriminate expressions.

\noindent\textbf{MAttNet} \cite{yu2018mattnet} is one of the most popular models for REF. In comparison with CMN which decomposes expression with fixed template, MAttNet propose a generic modular network with three modules for subject, location and relationship to address all kinds of referring expressions.

\noindent\textbf{VC} \cite{niu2019variational} is a recent state-of-the-art based on variational Beyesian method, called Variational Context (VC), to exploit the reciprocal relation between the referent and context.

\noindent\textbf{LGARNs} \cite{wang2019neighbourhood} is a graph-based reasoning model for referring expression. By building a directed graph over objects in an image and a language-guided graph attention network to highlight the relevant content in the expression, the model can explore the relationships between objects and make the grounding process explainable.

\noindent \textbf{Human.} We also test the human performance. In order to reduce inter-human variability, three workers are asked to choose the target object from the candidates, given the referring expression. If at least two of them selected the correct one,
then it is regarded as a success.

\begin{figure*}[htbp]
	\setlength{\abovecaptionskip}{0.3cm}
	\centering
	\includegraphics[width=0.85\textwidth]{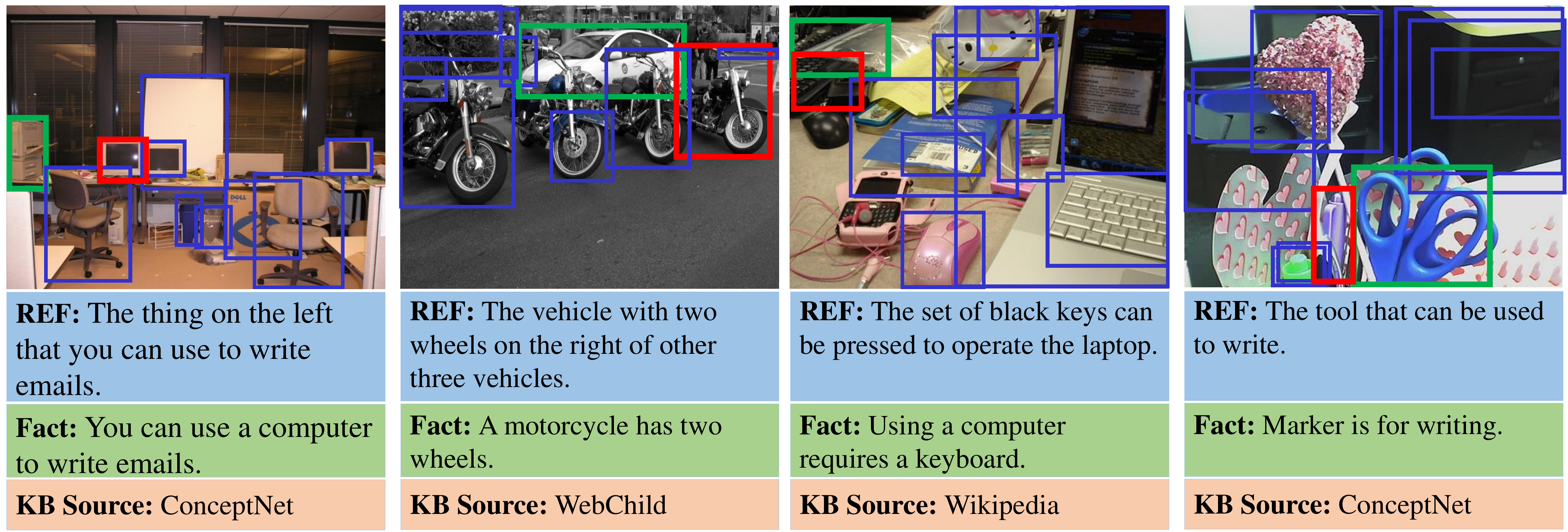}
	\caption{Visualization of some experimental results. Candidate bounding boxes (in blue) are presented in the image. The red one is the object selected by our algorithm. And the green one is chosen by the MAttNet.}
	\label{visualization}
	\vspace{-3mm}
\end{figure*}

The overall accuracy of all evaluated models with ground truth candidate objects bounding boxes are presented in Table~\ref{table:accuracy_gt}. All the SoTA models show a significant performance drop compared to their performance on RefCOCOs, {{which demonstrates the challenge of our dataset.}} Our model reaches an accuracy of $58.97 \%$ on KB-Ref test set, outperforming all the SoTA models by nearly $12 \%$, which suggests the necessity of exploring external knowledge in our REF setting. In addition, there is still a large gap between our model and the human performance (about $30 \%$ in accuracy). We also visualize some experimental results on Figure~\ref{visualization}.
We also add the proposed episodic memory module (EMM) into MAttNet, 
which improves the test accuracy from $46.03\%$ to $63.57\%$.
It further validates the importance of commonsense knowledge integration for our proposed REF task and the effectiveness of EMM.

\begin{table}
	{
		\begin{center}
			\scalebox{0.8}{
				\begin{tabular}{l|c|c}
					\hline
					Method & \multicolumn{2}{|c}{Accuracy (\%)}
					\\ \cline{2-3}	&  Val & Test \\
					\hline
					CMN \cite{hu2017modeling} & $41.28$ & $40.03$ \\
					SLR \cite{yu2017joint} & $44.03$ & $42.92$ \\
					VC \cite{niu2019variational} & $44.63$ & $43.59$ \\
					LGARNs \cite{wang2019neighbourhood} & $45.11$ &$44.27$ \\
					MAttNet \cite{yu2018mattnet} & $46.86$ & $46.03$ \\
					\hline
					\modelname~(Ours) & $59.45$ & $58.97$ \\
					MAttNet \cite{yu2018mattnet} + EMM & $64.08$ & $63.57$\\
					\hline
					Human performance & - & $90.13$ \\
					\hline
				\end{tabular}
			}
		\end{center}
	}
	\caption{Performance (Acc\%) comparison with SoTA REF approaches and our proposed \modelname~ on \sexyname. Our \modelname~ shows the highest accuracy on both validation and test set. All listed models use VGG-16 features.}
	\label{table:accuracy_gt}
	\vspace{-0.7cm}
\end{table}

\subsection{Ablation Studies}
%Next, we conduct ablation studies to analyze the dataset and investigate the key components of our \modelname.  

%\subsubsection{Effectiveness of Dynamic Memory Network}

%To validate the effectiveness and universality of the proposed dynamic memory network, we incorporate it into the popular MAttNet. Besides the subject, location and relationship phase parsing in the given expression, commonsense knowledge part is also softly decomposed and embedded by the language attention module, which is then used to trigger the additional dynamic memory network as designed in Section~\ref{sec:dmn}. %to compute a matching score. 
%An overall score is then computed as a weighted combination of four module scores. The whole model is trained with the same training strategy as original MAttNet. Results in Table~\ref{table:abalation study} show that, with the usage of dynamic memory network, the accuracy consistently increase about $- \%$ on validation and test, which indicates the effectiveness of the proposed fact attention model.
%on commonsense fact integration. 
%The accuracy is even higher than our \modelname, which, we believe, comes from the elaborately designed modular attention networks.

\begin{table}
	{
		\begin{center}
			\scalebox{0.8}{	
				\begin{tabular}{l|c|c}
					\hline
					Method & \multicolumn{2}{|c}{Accuracy (\%)}
					\\ \cline{2-3}	&  Val & Test \\
					\hline
					\modelname~ (Soft Attention) & $56.03$ & $54.92$ \\
					\hline
					\modelname~(EMM, $1$-pass) & $57.49$ & $56.57$ \\
					\modelname~(EMM, $3$-pass) & $58.52$ & $57.83$ \\
					\modelname~(EMM, $5$-pass) & $59.45$ & $58.97$ \\
					\modelname~(EMM, $10$-pass) & $59.47$ & $58.99$ \\
					\hline
					\modelname~(EMM, $5$-pass) + Facts Supervision & $59.50$ & $59.01$ \\
					\modelname~(EMM, $5$-pass)-ResNet-50 & $60.60$ & $60.35$ \\
					\hline
				\end{tabular}
			}
		\end{center}
	}
	\caption{Ablation study on the facts attention module. 
		Both using $1$ pass, our episodic memory module (EMM) with an attentional LSTM performs better than the naive soft attention, as the former considers the interaction between facts.
		The EMM performs better when the number of passes increases, which shows the effectiveness of the multi-hop attention strategy used by EMM.
		The extra fact supervision leads to a marginal accuracy increase. 
	}
	\label{table:abalation_study}
	\vspace{-0.9cm}
\end{table}

\begin{figure*}[htbp]
	\setlength{\abovecaptionskip}{0.3cm}
	\centering
	\includegraphics[width=0.85\textwidth]{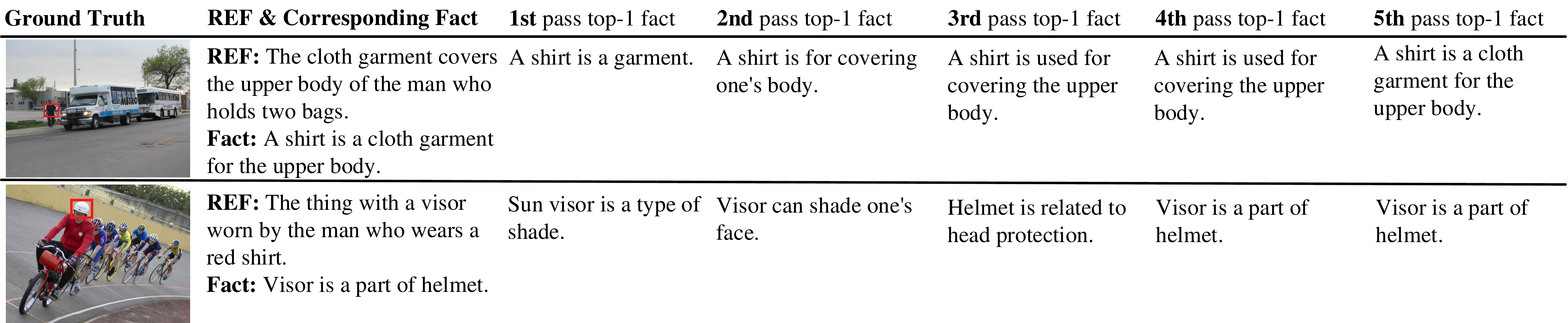}
	\caption{Visualization of the weight calculated by the Two-stage Fact Attention Module. We show the top-$1$ fact of each pass of the \modelname.}
	\label{mn_attention}
	\vspace{-4mm}
\end{figure*}

\paragraph{Effectiveness of Episodic Memory Module}
The adopted Episodic Memory Module (EMM) performs a multi-hop attention process.
To validate it effectiveness, we compare it with a single-pass soft attention module and also evaluate it with different numbers of passes $T = 1, 3, 5, 10$.
The single-pass ($T=1$) soft attention module compute the attended facts feature using a weighted sum $\mathbf{f}_n^e = \sum_{k=1}^K \alpha_k \mathbf{s}_k $, instead of using the attentional LSTM as Equation~\ref{mnm_2}.
It does not take into consideration the interaction between facts. 
As shown in Table~\ref{table:abalation_study}, the episodic memory module with one pass already surpasses soft attention by roughly $2.4$ percentages.
We also observe that the accuracy is improved with the increase of the number of passes $T$, which validates the advantage of multi-hop attention.
As the performance almost saturates at $T=5$, 
we choose the model \modelname~(EMM, $5$-pass) for the following experiments,
in order to strike a balance between accuracy and speed.
In Figure~\ref{mn_attention}, we also showcase the focused facts over different passes.

\paragraph{Impact of Direct Facts Supervision}
Note that our episodic memory module is trained in a weakly supervised manner by the remote cross entropy loss for object selection.
As the groundtruth supporting fact has been recorded in our dataset, 
it is straightforward to add a direct supervision on the episodic memory module.
To be specific, a target vector is defined where the position corresponding to the ground truth fact is filled with $1$ while others are $0$. A binary cross entropy function is then employed to calculate the loss between the facts attention weights and the target vector, which is applied on the last pass of EMM.
The corresponding results in Table~\ref{table:abalation_study} shows that 
adding direct fact supervision does not yield significantly better performance, which means that the weak supervision is considerably strong for training episodic memory module. 

\paragraph{Impact of Image feature Extractor}
We also try to use ResNet-50 to extract the image feature.
As shown in Table~\ref{table:abalation_study}, with replacing the VGG-16 by ResNet-50~\cite{he2016deep}, the performance increases by $1.3$ percentages.

\begin{table}[tb]
	{
		\begin{center}
			\scalebox{0.8}{
				\begin{tabular}{l|c|c}
					\hline
					Method & \multicolumn{2}{|c}{Accuracy (\%)}
					\\ \cline{2-3}	&  Val & Test \\
					\hline
					CMN \cite{hu2017modeling} & $20.91$ & $20.46$ \\
					SLR \cite{yu2017joint} & $21.33$ & $20.92$ \\
					VC \cite{niu2019variational} & $21.68$ & $21.29$ \\
					LGARNs \cite{wang2019neighbourhood} & $21.72$ & $21.37$ \\
					MAttNet \cite{yu2018mattnet} & $22.04$ & $21.73$ \\
					\hline
					%\modelname~ (no facts) & $19.46$ & $20.09$ \\
					\modelname (ours)  & $24.11$ & $23.82$ \\
					\hline
					
				\end{tabular}
			}
		\end{center}
	}
	\caption{Performance (Acc\%) comparison with SoTA REF approaches and our proposed \modelname~ on KB-Ref, using detected bounding boxes.} 
	\label{table:accuracy}
	\vspace{-1cm}
\end{table}

\paragraph{Comparison with SOTA using Detected Bounding Boxes}
We also evaluate the performance with detected bounding boxes. 
A $1600$-category Faster-RCNN detector is trained on Visual Genome and then applied on the validation and test images to extract object bounding box and category. The top-$10$ scored bounding boxes are extracted for each image. From Table~\ref{table:accuracy}, we can see that the gap between our model and other counterparts that without using knowledge is roughly $2 \sim 3\%$, which is significantly smaller than the gap achieved using ground-truth bounding boxes. The reason is that the trained detector is far from satisfactory, which only generates low-quality and misleading bounding boxes and labels. A wrong label may lead our proposed model to extract incorrect knowledge from the knowledge base.

%% file: conclusion.tex
\section{Conclusion}
In this work, we present a referring expression dataset, \sexyname, in which the objects are expressed by their visual and non-visual attributes. Such that, it encourages REF algorithms to explore information from images as well as external knowledge bases. The dataset features a large variety of objects ($1,805$ categories) and long expressions ($13.32$ in average).
Due to its complexity, directly applying SoTA REF approaches does not achieve promising results. To this end, we propose to tackle the problem with a 
expression conditioned image and fact attention network (\modelname).
Experiments show that our proposed model indeed improves the performance on \sexyname by a large margin.

%% file: main.bbl

\begin{thebibliography}{42}

%%% the Web address.

\ifx \showCODEN    \undefined \def \showCODEN     #1{\unskip}     \fi
\ifx \showDOI      \undefined \def \showDOI       #1{#1}\fi
\ifx \showISBNx    \undefined \def \showISBNx     #1{\unskip}     \fi
\ifx \showISBNxiii \undefined \def \showISBNxiii  #1{\unskip}     \fi
\ifx \showISSN     \undefined \def \showISSN      #1{\unskip}     \fi
\ifx \showLCCN     \undefined \def \showLCCN      #1{\unskip}     \fi
\ifx \shownote     \undefined \def \shownote      #1{#1}          \fi
\ifx \showarticletitle \undefined \def \showarticletitle #1{#1}   \fi
\ifx \showURL      \undefined \def \showURL       {\relax}        \fi
% invisible to TeX
\providecommand\bibfield[2]{#2}
\providecommand\bibinfo[2]{#2}
\providecommand\natexlab[1]{#1}
\providecommand\showeprint[2][]{arXiv:#2}

\bibitem[\protect\citeauthoryear{Cirik, Morency, and Berg-Kirkpatrick}{Cirik
  et~al\mbox{.}}{2018}]%
        {cirik2018visual}
\bibfield{author}{\bibinfo{person}{Volkan Cirik},
  \bibinfo{person}{Louis-Philippe Morency}, {and} \bibinfo{person}{Taylor
  Berg-Kirkpatrick}.} \bibinfo{year}{2018}\natexlab{}.
\newblock \showarticletitle{Visual Referring Expression Recognition: What Do
  Systems Actually Learn?}
\newblock \bibinfo{journal}{\emph{arXiv preprint arXiv:1805.11818}}
  (\bibinfo{year}{2018}).
\newblock


\bibitem[\protect\citeauthoryear{de~Vries, Strub, Chandar, Pietquin,
  Larochelle, and Courville}{de~Vries et~al\mbox{.}}{2017}]%
        {GuessWhat}
\bibfield{author}{\bibinfo{person}{Harm de Vries}, \bibinfo{person}{Florian
  Strub}, \bibinfo{person}{Sarath Chandar}, \bibinfo{person}{Olivier Pietquin},
  \bibinfo{person}{Hugo Larochelle}, {and} \bibinfo{person}{Aaron Courville}.}
  \bibinfo{year}{2017}\natexlab{}.
\newblock \showarticletitle{GuessWhat?!: Visual object discovery through
  multi-modal dialogue}. In \bibinfo{booktitle}{\emph{{Proc. IEEE Conf. Comp.
  Vis. Patt. Recogn.}}}
\newblock


\bibitem[\protect\citeauthoryear{Deng, Wu, Wu, Hu, Lyu, and Tan}{Deng
  et~al\mbox{.}}{2018}]%
        {Chaoruicvpr2018}
\bibfield{author}{\bibinfo{person}{Chaorui Deng}, \bibinfo{person}{Qi Wu},
  \bibinfo{person}{Qingyao Wu}, \bibinfo{person}{Fuyuan Hu},
  \bibinfo{person}{Fan Lyu}, {and} \bibinfo{person}{Mingkui Tan}.}
  \bibinfo{year}{2018}\natexlab{}.
\newblock \showarticletitle{Visual Grounding via Accumulated Attention}. In
  \bibinfo{booktitle}{\emph{{Proc. IEEE Conf. Comp. Vis. Patt. Recogn.}}}
\newblock


\bibitem[\protect\citeauthoryear{Gu, Zhao, Lin, Li, Cai, and Ling}{Gu
  et~al\mbox{.}}{2019}]%
        {gu2019scene}
\bibfield{author}{\bibinfo{person}{Jiuxiang Gu}, \bibinfo{person}{Handong
  Zhao}, \bibinfo{person}{Zhe Lin}, \bibinfo{person}{Sheng Li},
  \bibinfo{person}{Jianfei Cai}, {and} \bibinfo{person}{Mingyang Ling}.}
  \bibinfo{year}{2019}\natexlab{}.
\newblock \showarticletitle{Scene graph generation with external knowledge and
  image reconstruction}. In \bibinfo{booktitle}{\emph{{Proc. IEEE Conf. Comp.
  Vis. Patt. Recogn.}}} \bibinfo{pages}{1969--1978}.
\newblock


\bibitem[\protect\citeauthoryear{He, Zhang, Ren, and Sun}{He
  et~al\mbox{.}}{2016}]%
        {he2016deep}
\bibfield{author}{\bibinfo{person}{Kaiming He}, \bibinfo{person}{Xiangyu
  Zhang}, \bibinfo{person}{Shaoqing Ren}, {and} \bibinfo{person}{Jian Sun}.}
  \bibinfo{year}{2016}\natexlab{}.
\newblock \showarticletitle{Deep residual learning for image recognition}. In
  \bibinfo{booktitle}{\emph{{Proc. IEEE Conf. Comp. Vis. Patt. Recogn.}}}
  \bibinfo{pages}{770--778}.
\newblock


\bibitem[\protect\citeauthoryear{Hu, Rohrbach, Andreas, Darrell, and Saenko}{Hu
  et~al\mbox{.}}{2017}]%
        {hu2017modeling}
\bibfield{author}{\bibinfo{person}{Ronghang Hu}, \bibinfo{person}{Marcus
  Rohrbach}, \bibinfo{person}{Jacob Andreas}, \bibinfo{person}{Trevor Darrell},
  {and} \bibinfo{person}{Kate Saenko}.} \bibinfo{year}{2017}\natexlab{}.
\newblock \showarticletitle{Modeling relationships in referential expressions
  with compositional modular networks}. In \bibinfo{booktitle}{\emph{{Proc.
  IEEE Conf. Comp. Vis. Patt. Recogn.}}} \bibinfo{pages}{1115--1124}.
\newblock


\bibitem[\protect\citeauthoryear{Hu, Xu, Rohrbach, Feng, Saenko, and
  Darrell}{Hu et~al\mbox{.}}{2016}]%
        {Hu2016cvpr}
\bibfield{author}{\bibinfo{person}{Ronghang Hu}, \bibinfo{person}{Huazhe Xu},
  \bibinfo{person}{Marcus Rohrbach}, \bibinfo{person}{Jiashi Feng},
  \bibinfo{person}{Kate Saenko}, {and} \bibinfo{person}{Trevor Darrell}.}
  \bibinfo{year}{2016}\natexlab{}.
\newblock \showarticletitle{Natural language object retrieval}. In
  \bibinfo{booktitle}{\emph{{Proc. IEEE Conf. Comp. Vis. Patt. Recogn.}}}
\newblock


\bibitem[\protect\citeauthoryear{Hudson and Manning}{Hudson and
  Manning}{2019}]%
        {hudson2019gqa}
\bibfield{author}{\bibinfo{person}{Drew~A Hudson} {and}
  \bibinfo{person}{Christopher~D Manning}.} \bibinfo{year}{2019}\natexlab{}.
\newblock \showarticletitle{Gqa: A new dataset for real-world visual reasoning
  and compositional question answering}. In \bibinfo{booktitle}{\emph{{Proc.
  IEEE Conf. Comp. Vis. Patt. Recogn.}}} \bibinfo{pages}{6700--6709}.
\newblock


\bibitem[\protect\citeauthoryear{Kazemzadeh, Ordonez, Matten, and
  Berg}{Kazemzadeh et~al\mbox{.}}{2014}]%
        {kazemzadeh2014referitgame}
\bibfield{author}{\bibinfo{person}{Sahar Kazemzadeh}, \bibinfo{person}{Vicente
  Ordonez}, \bibinfo{person}{Mark Matten}, {and} \bibinfo{person}{Tamara
  Berg}.} \bibinfo{year}{2014}\natexlab{}.
\newblock \showarticletitle{Referitgame: Referring to objects in photographs of
  natural scenes}. In \bibinfo{booktitle}{\emph{{Proc. Conf. Empirical Methods
  in Natural Language Processing}}}. \bibinfo{pages}{787--798}.
\newblock


\bibitem[\protect\citeauthoryear{Krishna, Zhu, Groth, Johnson, Hata, Kravitz,
  Chen, Kalantidis, Li, Shamma, et~al\mbox{.}}{Krishna et~al\mbox{.}}{2017}]%
        {krishna2017visual}
\bibfield{author}{\bibinfo{person}{Ranjay Krishna}, \bibinfo{person}{Yuke Zhu},
  \bibinfo{person}{Oliver Groth}, \bibinfo{person}{Justin Johnson},
  \bibinfo{person}{Kenji Hata}, \bibinfo{person}{Joshua Kravitz},
  \bibinfo{person}{Stephanie Chen}, \bibinfo{person}{Yannis Kalantidis},
  \bibinfo{person}{Li-Jia Li}, \bibinfo{person}{David~A Shamma},
  {et~al\mbox{.}}} \bibinfo{year}{2017}\natexlab{}.
\newblock \showarticletitle{Visual genome: Connecting language and vision using
  crowdsourced dense image annotations}.
\newblock \bibinfo{journal}{\emph{{Int. J. Comput. Vision}}}
  \bibinfo{volume}{123}, \bibinfo{number}{1} (\bibinfo{year}{2017}),
  \bibinfo{pages}{32--73}.
\newblock


\bibitem[\protect\citeauthoryear{Lee, Fang, Yeh, and Frank~Wang}{Lee
  et~al\mbox{.}}{2018}]%
        {lee2018multi}
\bibfield{author}{\bibinfo{person}{Chung-Wei Lee}, \bibinfo{person}{Wei Fang},
  \bibinfo{person}{Chih-Kuan Yeh}, {and} \bibinfo{person}{Yu-Chiang
  Frank~Wang}.} \bibinfo{year}{2018}\natexlab{}.
\newblock \showarticletitle{Multi-label zero-shot learning with structured
  knowledge graphs}. In \bibinfo{booktitle}{\emph{{Proc. IEEE Conf. Comp. Vis.
  Patt. Recogn.}}} \bibinfo{pages}{1576--1585}.
\newblock


\bibitem[\protect\citeauthoryear{Li, Su, and Zhu}{Li et~al\mbox{.}}{2018}]%
        {KDMN}
\bibfield{author}{\bibinfo{person}{Guohao Li}, \bibinfo{person}{Hang Su}, {and}
  \bibinfo{person}{Wenwu Zhu}.} \bibinfo{year}{2018}\natexlab{}.
\newblock \showarticletitle{Incorporating External Knowledge to Answer
  Open-Domain Visual Questionswith Dynamic Memory Networks}. In
  \bibinfo{booktitle}{\emph{{Proc. IEEE Conf. Comp. Vis. Patt. Recogn.}}}
\newblock


\bibitem[\protect\citeauthoryear{Li, Wang, Shen, and van~den Hengel}{Li
  et~al\mbox{.}}{2019}]%
        {licvpr2019}
\bibfield{author}{\bibinfo{person}{Hui Li}, \bibinfo{person}{Peng Wang},
  \bibinfo{person}{Chunhua Shen}, {and} \bibinfo{person}{Anton van~den
  Hengel}.} \bibinfo{year}{2019}\natexlab{}.
\newblock \showarticletitle{Visual Question Answering as Reading
  Comprehension}. In \bibinfo{booktitle}{\emph{{Proc. IEEE Conf. Comp. Vis.
  Patt. Recogn.}}}
\newblock


\bibitem[\protect\citeauthoryear{Lin, Maire, Belongie, Hays, Perona, Ramanan,
  Dollar, and Zitnick}{Lin et~al\mbox{.}}{2014}]%
        {MSCOCO}
\bibfield{author}{\bibinfo{person}{Tsungyi Lin}, \bibinfo{person}{Michael
  Maire}, \bibinfo{person}{Serge Belongie}, \bibinfo{person}{James Hays},
  \bibinfo{person}{Pietro Perona}, \bibinfo{person}{Deva Ramanan},
  \bibinfo{person}{Piotr Dollar}, {and} \bibinfo{person}{C~Lawrencek Zitnick}.}
  \bibinfo{year}{2014}\natexlab{}.
\newblock \showarticletitle{Microsoft coco: Common objects in context.}. In
  \bibinfo{booktitle}{\emph{{Proc. Eur. Conf. Comp. Vis.}}}
\newblock


\bibitem[\protect\citeauthoryear{Liu, Zhang, Wu, and Zha}{Liu
  et~al\mbox{.}}{2019b}]%
        {DaqingICCV2019}
\bibfield{author}{\bibinfo{person}{Daqing Liu}, \bibinfo{person}{Hanwang
  Zhang}, \bibinfo{person}{Feng Wu}, {and} \bibinfo{person}{ZhengJun Zha}.}
  \bibinfo{year}{2019}\natexlab{b}.
\newblock \showarticletitle{Learning to Assemble Neural Module Tree Networks
  for Visual Grounding}. In \bibinfo{booktitle}{\emph{{Proc. IEEE Int. Conf.
  Comp. Vis.}}}
\newblock


\bibitem[\protect\citeauthoryear{Liu, Liu, Bai, and Yuille}{Liu
  et~al\mbox{.}}{2019a}]%
        {liu2019clevr}
\bibfield{author}{\bibinfo{person}{Runtao Liu}, \bibinfo{person}{Chenxi Liu},
  \bibinfo{person}{Yutong Bai}, {and} \bibinfo{person}{Alan~L Yuille}.}
  \bibinfo{year}{2019}\natexlab{a}.
\newblock \showarticletitle{Clevr-ref+: Diagnosing visual reasoning with
  referring expressions}. In \bibinfo{booktitle}{\emph{{Proc. IEEE Conf. Comp.
  Vis. Patt. Recogn.}}} \bibinfo{pages}{4185--4194}.
\newblock


\bibitem[\protect\citeauthoryear{Lu, Krishna, Bernstein, and Fei-Fei}{Lu
  et~al\mbox{.}}{2016}]%
        {lu2016visual}
\bibfield{author}{\bibinfo{person}{Cewu Lu}, \bibinfo{person}{Ranjay Krishna},
  \bibinfo{person}{Michael Bernstein}, {and} \bibinfo{person}{Li Fei-Fei}.}
  \bibinfo{year}{2016}\natexlab{}.
\newblock \showarticletitle{Visual relationship detection with language
  priors}. In \bibinfo{booktitle}{\emph{{Proc. Eur. Conf. Comp. Vis.}}}
  Springer, \bibinfo{pages}{852--869}.
\newblock


\bibitem[\protect\citeauthoryear{Luo and Shakhnarovich}{Luo and
  Shakhnarovich}{2017}]%
        {Luo2017}
\bibfield{author}{\bibinfo{person}{Ruotian Luo} {and} \bibinfo{person}{Gregory
  Shakhnarovich}.} \bibinfo{year}{2017}\natexlab{}.
\newblock \showarticletitle{Comprehension-guided referring expressions.}. In
  \bibinfo{booktitle}{\emph{{Proc. IEEE Conf. Comp. Vis. Patt. Recogn.}}}
\newblock


\bibitem[\protect\citeauthoryear{Mao, Huang, Toshev, Camburu, Yuille, and
  Murphy}{Mao et~al\mbox{.}}{2016}]%
        {mao2016generation}
\bibfield{author}{\bibinfo{person}{Junhua Mao}, \bibinfo{person}{Jonathan
  Huang}, \bibinfo{person}{Alexander Toshev}, \bibinfo{person}{Oana Camburu},
  \bibinfo{person}{Alan Yuille}, {and} \bibinfo{person}{Kevin Murphy}.}
  \bibinfo{year}{2016}\natexlab{}.
\newblock \showarticletitle{Generation and Comprehension of Unambiguous Object
  Descriptions}. In \bibinfo{booktitle}{\emph{{Proc. IEEE Conf. Comp. Vis.
  Patt. Recogn.}}}
\newblock


\bibitem[\protect\citeauthoryear{Marino, Rastegari, Farhadi, and
  Mottagh}{Marino et~al\mbox{.}}{2019}]%
        {OKVQA}
\bibfield{author}{\bibinfo{person}{Kenneth Marino}, \bibinfo{person}{Mohammad
  Rastegari}, \bibinfo{person}{Ali Farhadi}, {and} \bibinfo{person}{Roozbeh
  Mottagh}.} \bibinfo{year}{2019}\natexlab{}.
\newblock \showarticletitle{{OK-VQA}: A Visual Question Answering Benchmark
  RequiringExternal Knowledge}. In \bibinfo{booktitle}{\emph{{Proc. IEEE Conf.
  Comp. Vis. Patt. Recogn.}}}
\newblock


\bibitem[\protect\citeauthoryear{Mikolov, Sutskever, Chen, Corrado, and
  Dean}{Mikolov et~al\mbox{.}}{2013}]%
        {Mikolov2013Distributed}
\bibfield{author}{\bibinfo{person}{Tomas Mikolov}, \bibinfo{person}{Ilya
  Sutskever}, \bibinfo{person}{Kai Chen}, \bibinfo{person}{Greg~S Corrado},
  {and} \bibinfo{person}{Jeff Dean}.} \bibinfo{year}{2013}\natexlab{}.
\newblock \showarticletitle{Distributed representations of words and phrases
  and their compositionality}. In \bibinfo{booktitle}{\emph{Advances in neural
  information processing systems}}. \bibinfo{pages}{3111--3119}.
\newblock


\bibitem[\protect\citeauthoryear{Miller}{Miller}{1995}]%
        {miller1995wordnet}
\bibfield{author}{\bibinfo{person}{George~A Miller}.}
  \bibinfo{year}{1995}\natexlab{}.
\newblock \showarticletitle{WordNet: a lexical database for English}.
\newblock \bibinfo{journal}{\emph{Commun. ACM}} \bibinfo{volume}{38},
  \bibinfo{number}{11} (\bibinfo{year}{1995}), \bibinfo{pages}{39--41}.
\newblock


\bibitem[\protect\citeauthoryear{Narasimhan, Lazebnik, and Schwing}{Narasimhan
  et~al\mbox{.}}{2018}]%
        {Narasimhan2018}
\bibfield{author}{\bibinfo{person}{Medhini Narasimhan},
  \bibinfo{person}{Svetlana Lazebnik}, {and} \bibinfo{person}{Alexander~G.
  Schwing}.} \bibinfo{year}{2018}\natexlab{}.
\newblock \showarticletitle{Out of the box: Reasoning with graph convolution
  nets for factual visual question answering}. In
  \bibinfo{booktitle}{\emph{{Proc. Advances in Neural Inf. Process. Syst.}}}
  \bibinfo{pages}{2654--2665}.
\newblock


\bibitem[\protect\citeauthoryear{Narasimhan and Schwing}{Narasimhan and
  Schwing}{2018}]%
        {narasimhan2018straight}
\bibfield{author}{\bibinfo{person}{Medhini Narasimhan} {and}
  \bibinfo{person}{Alexander~G Schwing}.} \bibinfo{year}{2018}\natexlab{}.
\newblock \showarticletitle{Straight to the facts: Learning knowledge base
  retrieval for factual visual question answering}. In
  \bibinfo{booktitle}{\emph{{Proc. Eur. Conf. Comp. Vis.}}}
  \bibinfo{pages}{451--468}.
\newblock


\bibitem[\protect\citeauthoryear{Niu, Zhang, Lu, and Chang}{Niu
  et~al\mbox{.}}{2019}]%
        {niu2019variational}
\bibfield{author}{\bibinfo{person}{Yulei Niu}, \bibinfo{person}{Hanwang Zhang},
  \bibinfo{person}{Zhiwu Lu}, {and} \bibinfo{person}{Shih-Fu Chang}.}
  \bibinfo{year}{2019}\natexlab{}.
\newblock \showarticletitle{Variational Context: Exploiting Visual and Textual
  Context for Grounding Referring Expressions}.
\newblock \bibinfo{journal}{\emph{{{IEEE} Trans. Pattern Anal. Mach. Intell.}}}
  (\bibinfo{year}{2019}).
\newblock


\bibitem[\protect\citeauthoryear{Rohrbach, Rohrbach, Hu, Darrell, and
  Schiele}{Rohrbach et~al\mbox{.}}{2016}]%
        {GroundeR}
\bibfield{author}{\bibinfo{person}{Anna Rohrbach}, \bibinfo{person}{Marcus
  Rohrbach}, \bibinfo{person}{Ronghang Hu}, \bibinfo{person}{Trevor Darrell},
  {and} \bibinfo{person}{Bernt Schiele}.} \bibinfo{year}{2016}\natexlab{}.
\newblock \showarticletitle{Grounding of Textual Phrases in Images by
  Reconstruction}. In \bibinfo{booktitle}{\emph{{Proc. Eur. Conf. Comp. Vis.}}}
  \bibinfo{pages}{817--834}.
\newblock


\bibitem[\protect\citeauthoryear{Speer, Chin, and Havasi}{Speer
  et~al\mbox{.}}{2017}]%
        {speer2017conceptnet}
\bibfield{author}{\bibinfo{person}{Robyn Speer}, \bibinfo{person}{Joshua Chin},
  {and} \bibinfo{person}{Catherine Havasi}.} \bibinfo{year}{2017}\natexlab{}.
\newblock \showarticletitle{Conceptnet 5.5: An open multilingual graph of
  general knowledge}. In \bibinfo{booktitle}{\emph{Thirty-First AAAI Conference
  on Artificial Intelligence}}.
\newblock


\bibitem[\protect\citeauthoryear{Su, Zhu, Dong, Cai, Chen, and Li}{Su
  et~al\mbox{.}}{2018}]%
        {VKMN}
\bibfield{author}{\bibinfo{person}{Zhou Su}, \bibinfo{person}{Chen Zhu},
  \bibinfo{person}{Yinpeng Dong}, \bibinfo{person}{Dongqi Cai},
  \bibinfo{person}{Yurong Chen}, {and} \bibinfo{person}{Jianguo Li}.}
  \bibinfo{year}{2018}\natexlab{}.
\newblock \showarticletitle{Learning Visual Knowledge Memory Networks for
  Visual Question Answering}. In \bibinfo{booktitle}{\emph{{Proc. IEEE Conf.
  Comp. Vis. Patt. Recogn.}}}
\newblock


\bibitem[\protect\citeauthoryear{Tandon, Melo, and Weikum}{Tandon
  et~al\mbox{.}}{2017}]%
        {Tandon2017WebChild}
\bibfield{author}{\bibinfo{person}{Niket Tandon}, \bibinfo{person}{Gerard~De
  Melo}, {and} \bibinfo{person}{Gerhard Weikum}.}
  \bibinfo{year}{2017}\natexlab{}.
\newblock \showarticletitle{WebChild 2.0 : Fine-Grained Commonsense Knowledge
  Distillation}. In \bibinfo{booktitle}{\emph{Proceedings of ACL 2017, System
  Demonstrations}}.
\newblock


\bibitem[\protect\citeauthoryear{Wang, Wu, Cao, Shen, Gao, and Hengel}{Wang
  et~al\mbox{.}}{2019}]%
        {wang2019neighbourhood}
\bibfield{author}{\bibinfo{person}{Peng Wang}, \bibinfo{person}{Qi Wu},
  \bibinfo{person}{Jiewei Cao}, \bibinfo{person}{Chunhua Shen},
  \bibinfo{person}{Lianli Gao}, {and} \bibinfo{person}{Anton van den~Hengel
  Hengel}.} \bibinfo{year}{2019}\natexlab{}.
\newblock \showarticletitle{Neighbourhood watch: Referring expression
  comprehension via language-guided graph attention networks}. In
  \bibinfo{booktitle}{\emph{{Proc. IEEE Conf. Comp. Vis. Patt. Recogn.}}}
  \bibinfo{pages}{1960--1968}.
\newblock


\bibitem[\protect\citeauthoryear{Wang, Wu, Shen, Dick, and van~den Hengel}{Wang
  et~al\mbox{.}}{2018a}]%
        {FVQA}
\bibfield{author}{\bibinfo{person}{Peng Wang}, \bibinfo{person}{Qi Wu},
  \bibinfo{person}{Chunhua Shen}, \bibinfo{person}{Anthony Dick}, {and}
  \bibinfo{person}{Anton van~den Hengel}.} \bibinfo{year}{2018}\natexlab{a}.
\newblock \showarticletitle{Fvqa: Fact-based visual question answering}.
\newblock \bibinfo{journal}{\emph{{{IEEE} Trans. Pattern Anal. Mach. Intell.}}}
  \bibinfo{volume}{40}, \bibinfo{number}{10} (\bibinfo{year}{2018}),
  \bibinfo{pages}{2413--2427}.
\newblock


\bibitem[\protect\citeauthoryear{Wang, Ye, and Gupta}{Wang
  et~al\mbox{.}}{2018b}]%
        {wang2018zero}
\bibfield{author}{\bibinfo{person}{Xiaolong Wang}, \bibinfo{person}{Yufei Ye},
  {and} \bibinfo{person}{Abhinav Gupta}.} \bibinfo{year}{2018}\natexlab{b}.
\newblock \showarticletitle{Zero-shot recognition via semantic embeddings and
  knowledge graphs}. In \bibinfo{booktitle}{\emph{{Proc. IEEE Conf. Comp. Vis.
  Patt. Recogn.}}} \bibinfo{pages}{6857--6866}.
\newblock


\bibitem[\protect\citeauthoryear{Wu, Wang, Shen, van~den Hengel, and Dick}{Wu
  et~al\mbox{.}}{2016}]%
        {askmeanything}
\bibfield{author}{\bibinfo{person}{Qi Wu}, \bibinfo{person}{Peng Wang},
  \bibinfo{person}{Chunhua Shen}, \bibinfo{person}{Anton van~den Hengel}, {and}
  \bibinfo{person}{Anthony Dick}.} \bibinfo{year}{2016}\natexlab{}.
\newblock \showarticletitle{Ask me anything: Free-form visual question
  answer-ing based on knowledge from external sources.}. In
  \bibinfo{booktitle}{\emph{{Proc. IEEE Conf. Comp. Vis. Patt. Recogn.}}}
\newblock


\bibitem[\protect\citeauthoryear{Xiong, Merity, and Socher}{Xiong
  et~al\mbox{.}}{2016}]%
        {Xiong2016Dynamic}
\bibfield{author}{\bibinfo{person}{Caiming Xiong}, \bibinfo{person}{Stephen
  Merity}, {and} \bibinfo{person}{Richard Socher}.}
  \bibinfo{year}{2016}\natexlab{}.
\newblock \showarticletitle{Dynamic memory networks for visual and textual
  question answering}. In \bibinfo{booktitle}{\emph{International conference on
  machine learning}}. \bibinfo{pages}{2397--2406}.
\newblock


\bibitem[\protect\citeauthoryear{Xu, Wong, Li, Zhao, and Kankanhalli}{Xu
  et~al\mbox{.}}{2019}]%
        {xu2019learning}
\bibfield{author}{\bibinfo{person}{Bingjie Xu}, \bibinfo{person}{Yongkang
  Wong}, \bibinfo{person}{Junnan Li}, \bibinfo{person}{Qi Zhao}, {and}
  \bibinfo{person}{Mohan~S Kankanhalli}.} \bibinfo{year}{2019}\natexlab{}.
\newblock \showarticletitle{Learning to detect human-object interactions with
  knowledge}. In \bibinfo{booktitle}{\emph{{Proc. IEEE Conf. Comp. Vis. Patt.
  Recogn.}}}
\newblock


\bibitem[\protect\citeauthoryear{Yang, Li, and Yu}{Yang et~al\mbox{.}}{2019}]%
        {SibeiICCV2019}
\bibfield{author}{\bibinfo{person}{Sibei Yang}, \bibinfo{person}{Guanbin Li},
  {and} \bibinfo{person}{Yizhou Yu}.} \bibinfo{year}{2019}\natexlab{}.
\newblock \showarticletitle{Dynamic Graph Attention for Referring Expression
  Comprehension}. In \bibinfo{booktitle}{\emph{{Proc. IEEE Int. Conf. Comp.
  Vis.}}}
\newblock


\bibitem[\protect\citeauthoryear{Yu, Lin, Shen, Yang, Lu, Bansal, and Berg}{Yu
  et~al\mbox{.}}{2018}]%
        {yu2018mattnet}
\bibfield{author}{\bibinfo{person}{Licheng Yu}, \bibinfo{person}{Zhe Lin},
  \bibinfo{person}{Xiaohui Shen}, \bibinfo{person}{Jimei Yang},
  \bibinfo{person}{Xin Lu}, \bibinfo{person}{Mohit Bansal}, {and}
  \bibinfo{person}{Tamara~L Berg}.} \bibinfo{year}{2018}\natexlab{}.
\newblock \showarticletitle{Mattnet: Modular attention network for referring
  expression comprehension}. In \bibinfo{booktitle}{\emph{{Proc. IEEE Conf.
  Comp. Vis. Patt. Recogn.}}} \bibinfo{pages}{1307--1315}.
\newblock


\bibitem[\protect\citeauthoryear{Yu, Poirson, Yang, Berg, and Berg}{Yu
  et~al\mbox{.}}{2016}]%
        {yu2016modeling}
\bibfield{author}{\bibinfo{person}{Licheng Yu}, \bibinfo{person}{Patrick
  Poirson}, \bibinfo{person}{Shan Yang}, \bibinfo{person}{Alexander~C Berg},
  {and} \bibinfo{person}{Tamara~L Berg}.} \bibinfo{year}{2016}\natexlab{}.
\newblock \showarticletitle{Modeling context in referring expressions}. In
  \bibinfo{booktitle}{\emph{{Proc. Eur. Conf. Comp. Vis.}}}
  \bibinfo{pages}{69--85}.
\newblock


\bibitem[\protect\citeauthoryear{Yu, Tan, Bansal, and Berg}{Yu
  et~al\mbox{.}}{2017b}]%
        {yu2017joint}
\bibfield{author}{\bibinfo{person}{Licheng Yu}, \bibinfo{person}{Hao Tan},
  \bibinfo{person}{Mohit Bansal}, {and} \bibinfo{person}{Tamara~L Berg}.}
  \bibinfo{year}{2017}\natexlab{b}.
\newblock \showarticletitle{A joint speaker-listener-reinforcer model for
  referring expressions}. In \bibinfo{booktitle}{\emph{{Proc. IEEE Conf. Comp.
  Vis. Patt. Recogn.}}} \bibinfo{pages}{7282--7290}.
\newblock


\bibitem[\protect\citeauthoryear{Yu, Li, Morariu, and Davis}{Yu
  et~al\mbox{.}}{2017a}]%
        {yu2017visual}
\bibfield{author}{\bibinfo{person}{Ruichi Yu}, \bibinfo{person}{Ang Li},
  \bibinfo{person}{Vlad~I Morariu}, {and} \bibinfo{person}{Larry~S Davis}.}
  \bibinfo{year}{2017}\natexlab{a}.
\newblock \showarticletitle{Visual relationship detection with internal and
  external linguistic knowledge distillation}. In
  \bibinfo{booktitle}{\emph{{Proc. IEEE Int. Conf. Comp. Vis.}}}
  \bibinfo{pages}{1974--1982}.
\newblock


\bibitem[\protect\citeauthoryear{Zellers, Bisk, Farhadi, and Choi}{Zellers
  et~al\mbox{.}}{2019}]%
        {zellers2019recognition}
\bibfield{author}{\bibinfo{person}{Rowan Zellers}, \bibinfo{person}{Yonatan
  Bisk}, \bibinfo{person}{Ali Farhadi}, {and} \bibinfo{person}{Yejin Choi}.}
  \bibinfo{year}{2019}\natexlab{}.
\newblock \showarticletitle{From recognition to cognition: Visual commonsense
  reasoning}. In \bibinfo{booktitle}{\emph{{Proc. IEEE Conf. Comp. Vis. Patt.
  Recogn.}}} \bibinfo{pages}{6720--6731}.
\newblock


\bibitem[\protect\citeauthoryear{Zhuang, Wu, Shen, Reid, and van~den
  Hengel}{Zhuang et~al\mbox{.}}{2018}]%
        {Bohancvpr2018}
\bibfield{author}{\bibinfo{person}{Bohan Zhuang}, \bibinfo{person}{Qi Wu},
  \bibinfo{person}{Chunhua Shen}, \bibinfo{person}{Ian Reid}, {and}
  \bibinfo{person}{Anton van~den Hengel}.} \bibinfo{year}{2018}\natexlab{}.
\newblock \showarticletitle{Parallel Attention: A Unified Framework for Visual
  Object Discoverythrough Dialogs and Queries}. In
  \bibinfo{booktitle}{\emph{{Proc. IEEE Conf. Comp. Vis. Patt. Recogn.}}}
\newblock


\end{thebibliography}
